\PassOptionsToPackage{colorlinks=true,linkcolor=blue,citecolor=blue,urlcolor=blue}{hyperref}
\documentclass[numbers]{article}

\usepackage[preprint]{neurips_2025}

\usepackage[utf8]{inputenc} 
\usepackage[T1]{fontenc}    
\usepackage{hyperref}       
\usepackage{url}            
\usepackage{booktabs}       
\usepackage{amsfonts}       
\usepackage{nicefrac}       
\usepackage{microtype}      
\usepackage{xcolor}         

\usepackage{microtype}
\usepackage{graphicx}
\usepackage{subfigure}
\usepackage{booktabs} 

\usepackage[utf8]{inputenc}
\usepackage{array}
\usepackage{longtable}
\usepackage{adjustbox}

\usepackage{xcolor}
\usepackage{colortbl}

\usepackage{caption} 

\usepackage{tcolorbox}
\usepackage{courier} 

\usepackage{listings}

\lstdefinestyle{prompt}{
  basicstyle=\ttfamily\small,
  breaklines=true,
  columns=fullflexible,
  frame=single,
  rulecolor=\color{black!20},
  backgroundcolor=\color{black!3},
  numbers=none,
  numberstyle=\tiny\color{black!50},
  xleftmargin=2mm, framexleftmargin=2mm,
  showstringspaces=false
}

\title{An Evaluation Study of Hybrid Methods for Multilingual PII Detection}

\author{
\textbf{Harshit Rajgarhia}\textsuperscript{1},
\textbf{Suryam Gupta}\textsuperscript{1},
\textbf{Asif Shaik}\textsuperscript{1},
\textbf{Gulipalli Praveen Kumar}\textsuperscript{1},\\[2pt]
\textbf{Y Santhoshraj}\textsuperscript{1},
\textbf{Sanka Nithya Tanvy Nishitha}\textsuperscript{1},
\textbf{Abhishek Mukherji}\textsuperscript{1}\\[4pt]
\textsuperscript{1}Centific Global Solutions Inc.\\[4pt]
}

\begin{document}

\maketitle

\vspace{-6mm}
\begin{abstract}
\vspace{-3mm}
The detection of Personally Identifiable Information (PII) is critical for privacy compliance but remains challenging in low-resource languages due to linguistic diversity and limited annotated data. We present \textbf{RECAP}, a hybrid framework that combines deterministic regular expressions with context-aware large language models (LLMs) for scalable PII detection across 13 low-resource locales. \textbf{RECAP}'s modular design supports over 300 entity types without retraining, using a three-phase refinement pipeline for disambiguation and filtering. \textbf{Benchmarked} with \texttt{nervaluate}, our system outperforms fine-tuned NER models by 82\% and zero-shot LLMs by 17\% in weighted F1-score. This work offers a scalable and adaptable solution for efficient PII detection in compliance-focused applications.
\end{abstract}

\vspace{-5mm}
\section{Introduction}
\vspace{-3mm}




The exponential growth of user-generated content has created vast repositories where Personally Identifiable Information (PII) often remains exposed, posing significant privacy risks and compliance challenges~\citep{baazarvoice, bigid}. To address these threats, regulations like GDPR, HIPAA, and CCPA mandate strict safeguards and penalties for non-compliance \cite{gdpreu, szczepanska2025phi, oag}. Additionally, as Large Language Models (LLMs) are deployed into production systems, robust evaluation across the model becomes paramount, and is especially critical for sensitive tasks like PII detection, which demand high precision, reliability, and adaptability. Yet, existing PII annotation systems struggle with ambiguity, format variability, and scaling over low-resource languages.



In this work, we propose \textbf{RECAP} (\textbf{RE}gex and \textbf{C}ontext-\textbf{A}ware \textbf{P}rompting), that combines a regex-based deterministic solution with context-enriched LLM for PII detection, addressing key limitations of Named Entity Recognition systems, zero-shot LLMs, and rule-based methods.

\textbf{Core Challenges Addressed:} (1) \textbf{Low-Resource Performance Gap:} Existing entity recognition systems often perform poorly in low-resource locales \footnote{"Low-resource" locales have limited publicly available annotated data for training.} due to lack of annotated training data, limited linguistic resources, and the high computational cost associated with training models for each new language or domain; (2) \textbf{Scalability Bottleneck:} Pure regex methods lack semantic understanding, while transformer-based NER models suffer from limited PII type coverage. Standalone LLMs, though flexible, produce inconsistent outputs and are prone to hallucination; and (3) \textbf{Ambiguity and Variation:} PII entities exhibit both structural variation and semantic ambiguity across locales, making them difficult to classify accurately using traditional approaches, which often result in missed and conflicting labels, thereby reducing overall reliability.


\textbf{Key Contributions:} (1) To our knowledge, our work is the first to introduce a PII detection framework that spans across 13 diverse, low-resource locales, with support for 300+ PII types across six domains, without requiring any model training or fine-tuning; (2) Our novel solution combines deterministic regex patterns with context-aware LLMs to advance PII detection for low-resource locales; (3) \textbf{RECAP} implements a three-phased pipeline (multi-label disambiguation, span consolidation, contextual filtering) to systematically reduce ambiguity and false positives; (4) We present detailed \textbf{benchmarking} results using \texttt{nervaluate} evaluation framework \cite{batista2025nervaluate}, where \textbf{RECAP} outperforms the state-of-the-art baselines, NER by 82\% and LLM by 17\% (weighted F1-score).
\vspace{-4mm}
\section{Related Work}
\vspace{-3mm}

\paragraph{Multilingual \& Low-Resource PII Detection:} Most work focuses on high-resource languages like English. Exceptions include deep learning for Luganda PII~\citep{africano2024pii}, and few-shot cross-lingual methods for clinical texts~\citep{amin-etal-2022-shot}. Datasets from MultiCoNER~\citep{malmasi-etal-2022-multiconer}, ~\citet{ai4privacy2023, bigcode2023} offer broader coverage but still lack low-resource representation — a gap our work addresses.

\vspace{-3mm}
\paragraph{Regex-based Detection:} Regular expressions (Regex) have been employed for general NER tasks like automated resume parsing \citep{sougandh2023automated} and high-risk PII detection \citep{subramani2023detecting}. However, this approach suffers from high false positives and performs poorly on unstructured formats.

\vspace{-3mm}
\paragraph{Deep Learning Methods:} Models like CASSED use BERT for structured data~\citep{kuvzina2023cassed}, while DTL-PIIE employs transfer learning for social media text~\citep{liu2021automated}. BERT-based models show strong performance with balanced data~\citep{mainetti2025detecting} but struggle with multi-label entities and numeric false positives.

\vspace{-3mm}
\paragraph{LLM-based Approaches:} LLMs are used for both PII detection and synthetic data generation (e.g., SPY~\citep{savkin2025spy}, ProgGen~\citep{heng-etal-2024-proggen}). They perform well in domain-specific settings like education~\citep{singhal2024identifying, shen2025enhancing} and chemistry~\citep{zhang2025rapid}. Models like GPT-NER frame detection as a generation task~\citep{wang-etal-2025-gpt}, and strategic zero-shot prompting detects sensitive information across global contexts~\citep{anand2024sensitive}. However, these approaches often suffer from over-redaction and hallucination, misidentifying non-PII. 
\vspace{-4mm}

\section{Solution Architecture and Workflow}
\vspace{-3mm}

\begin{figure*}[t]
  \centering

  \begin{minipage}[b]{0.48\textwidth}
    \centering
    \includegraphics[width=\textwidth, keepaspectratio]%
      {"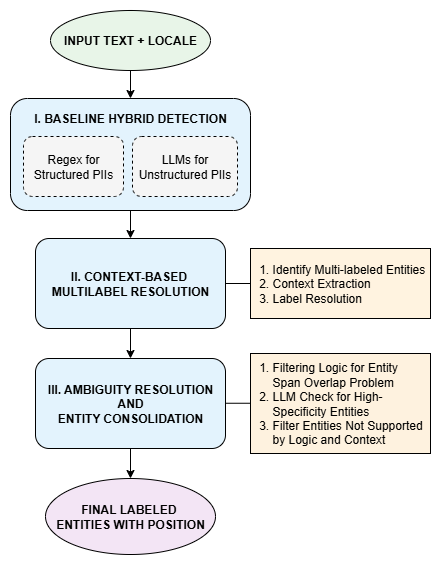"}
    \captionof{figure}{\textbf{RECAP} Architecture}
    \label{fig:workflow}
  \end{minipage}
  \hfill
  \begin{minipage}[b]{0.48\textwidth}
    \centering
    \includegraphics[width=\textwidth, keepaspectratio]{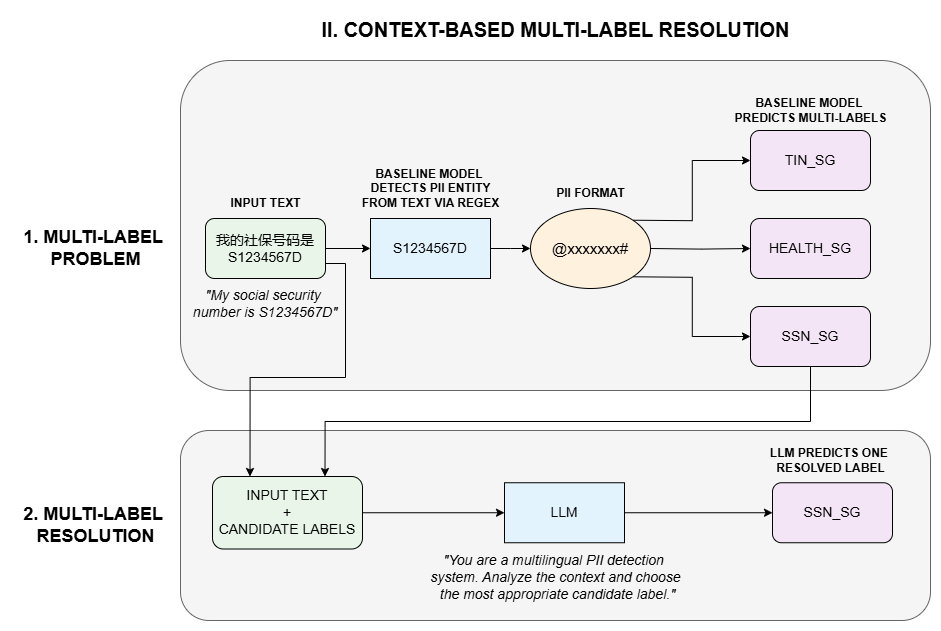}

    \vspace{1.5mm}

    \includegraphics[width=\textwidth, keepaspectratio]{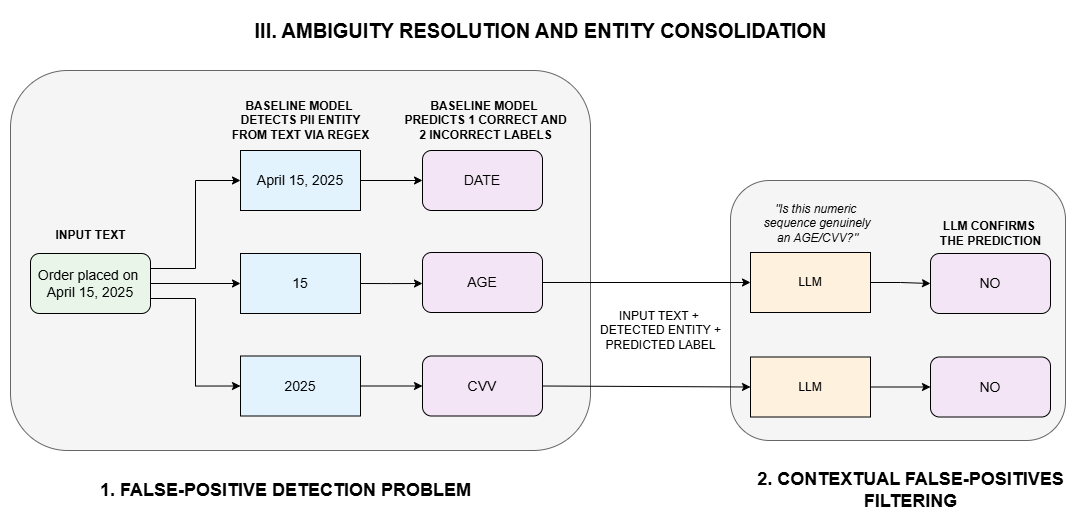}
    \captionof{figure}{Multi-labeling (top) and False Positives (bottom) detection problem and resolution}
    \label{fig:multi_FP}
  \end{minipage}
  \vspace{-5mm}

\end{figure*}

Our \textbf{RECAP} architecture (Figure~\ref{fig:workflow}) is designed to tackle the inherent challenges of multilingual PII detection by combining the precision of rule-based methods for structured PIIs \footnote{We refer to \textit{Structured PIIs} as those which have a syntactically regular, well-defined formats and are usually represented by numerical patterns, such as \texttt{SSN} (a unique ID number in the US). In a similar fashion, \textit{Unstructured PIIs} refers to entities which have semantically variable and arbitrary patterns, such as \texttt{ADDRESS}.}, with the semantic understanding of LLMs for unstructured PIIs. The system employs a modular, locale-aware design where each of the 13 supported locales has a dedicated detector containing its specific regex patterns and optimized prompts. The core of our approach is a three-phase refinement pipeline that progressively improves detection quality from an initial hybrid baseline to a final refined output, as summarized in Table~\ref{tab:pii_detection_phases_transposed} and evidenced by the performance gains across locales in Table~\ref{tab:locale_iterations}. Sample sizes with locale information are shown in Table~\ref{tab:locale_samples}.

\textbf{I. Baseline Hybrid Detection:} The process begins by receiving a text sample and its associated locale, invoking the corresponding locale-specific detector. The text is processed by a comprehensive set of regular expressions to detect structured PIIs. These can be categorized into two types: (a) universal patterns for entities like \texttt{IP\_ADDRESS} that follow global formats, and (b) locale-specific patterns for entities like national IDs (e.g., India's \texttt{AADHAAR\_IN} vs. Belgium's \texttt{SSN\_BE}), which require custom regex patterns per country. In parallel, the entire text is passed to an LLM (GPT-4o) using a carefully engineered zero-shot prompt (Listing~\ref{lst:name-prompt}) to detect unstructured PIIs (\texttt{NAME}, \texttt{ADDRESS}, \texttt{USERNAME}, and \texttt{PASSWORD}). This hybrid baseline provides broad coverage in terms of detecting PII entities, but introduces three key challenges: (1) \textbf{Multi-labeling} from semantically different but syntactically similar regex patterns; (2) \textbf{Span overlaps}, where one entity is fully or partially contained within another, leading to redundant and inconsistent labeling; and (3) \textbf{Contextual false positives} on short numeric sequences like \texttt{CVV} (Card Verification Value) or \texttt{AGE} that appear in non-sensitive contexts.

\textbf{II. Context-based Multi-label Resolution:} This phase focuses on resolving ambiguity in cases where a single entity span is assigned multiple candidate labels by the baseline. These multi-labeled outputs typically arise from identical syntactic patterns across numerically formatted entities. While the baseline regex method effectively identify entities, they may be unable to determine which label is most contextually appropriate. Our resolution module identifies all entities assigned multiple labels. For each, the original text, the character span, and the candidate labels are passed to the LLM with a custom prompt that instructs it to analyze the surrounding context and select the single most appropriate label (Figure~\ref{fig:multi_FP}, Top). This leverages the LLM's semantic understanding to resolve ambiguities that are intractable for rules alone and ensures consistent, context-aware labeling, significantly boosting precision and recall.

\textbf{III. Ambiguity Resolution and Entity Consolidation:} The final phase applies two targeted filters to produce a clean, coherent set of predictions. (1) \textbf{Entity Span Overlap Resolution}: A deterministic algorithm processes all entities sorted by their start position, pre-defined label priority, and span length. It removes entities that are fully contained within a longer span if they have a lower priority (e.g., filtering out \texttt{AGE}="24" when it is contained within a correctly identified \texttt{ADDRESS}="24 Lincoln Avenue, NY"). (2) \textbf{Contextual False Positive Filtering}: For high-specificity, short numeric entities (\texttt{AGE}, \texttt{CVV}), a local context window (one sentence before and after the entity) is extracted. This context is submitted to the LLM to verify whether the numeric value is semantically plausible as the predicted PII type. An entity is retained only upon LLM confirmation (Figure~\ref{fig:multi_FP}, Bottom), drastically reducing false positives that arise from numeric coincidences in non-PII settings, thereby improving overall precision while maintaining recall.

Table~\ref{tab:locale_iterations} and~\ref{tab:ner_iteration_results} highlights consistent F1-score improvements across the three phases of \textbf{RECAP} architecture. Most locales showed steady gains, with \textsc{sv\_SE} and \textsc{pt\_BR} achieving notable increases of 77.53\% and 47.76\%, respectively. While \textsc{nl\_BE} showed marginal fluctuations due to high initial performance, the overall trend highlights the efficacy of each refinement in enhancing PII detection.

\begin{table*}[htbp]
\captionsetup{skip=2pt}  
\centering
\vspace{-2mm}  
\caption{Performance progression across different phases}
\label{tab:pii_detection_phases_transposed}
\small
\setlength{\tabcolsep}{4pt}  
\renewcommand{\arraystretch}{1.05}  
\resizebox{\textwidth}{!}{%
\begin{tabular}{|p{1.7cm}|p{4.2cm}|p{4.2cm}|p{4.2cm}|}
\hline
\textbf{Phases} & \textbf{Baseline Hybrid Detection} & \textbf{Context Based Label Resolution} & \textbf{Ambiguity and Entity Consolidation} \\
\hline
\textbf{Solution Approach} & Regex detection for structured entities and zero-shot LLM for unstructured ones & Multi label disambiguation using LLM with context-aware resolution & Suppression of overlap spans and false positives using logic filtering and LLM check \\
\hline
\textbf{Weighted F1 Score (Gain)} 
& \raisebox{-1.1ex}{0.511 (-)} 
& \raisebox{-1.1ex}{0.585 ($\Delta \approx 14.48\%$)} 
& \raisebox{-1.1ex}{0.657 ($\Delta \approx 12.30\%$)} \\
\hline
\textbf{Impact} & Establishes initial coverage across 13 locales using structured rules and semantic generalization & Reduces incorrect or overly generic labels to improve precision and semantic alignment & Filters short numeric patterns and improves consistency by resolving labels in overlap spans \\
\hline
\end{tabular}
}
\vspace{-4mm}  
\end{table*}
\section{Benchmark Results and Comparative Evaluation}
\vspace{-3mm}


\textbf{Benchmark Design:} We evaluate \textbf{RECAP} against two strong baselines: (1) \textbf{transformers-based NER}: We select the best available HuggingFace model for each locale (see Appendix, Figure~\ref{fig:ner-model-codes}), representing traditional fine-tuned entity recognition models. These models are typically limited to generic labels (\texttt{PER}, \texttt{ORG}, \texttt{LOC}, \texttt{MISC}). (2) \textbf{Zero-shot LLM}: We use GPT-4o with a natural language prompt for PII extraction. While adaptable to multiple languages and label types, these models exhibit high variability in outputs, inconsistent formatting, and require careful prompt design.

\begin{figure*}[htbp]
    \centering
    \includegraphics[width=\textwidth]{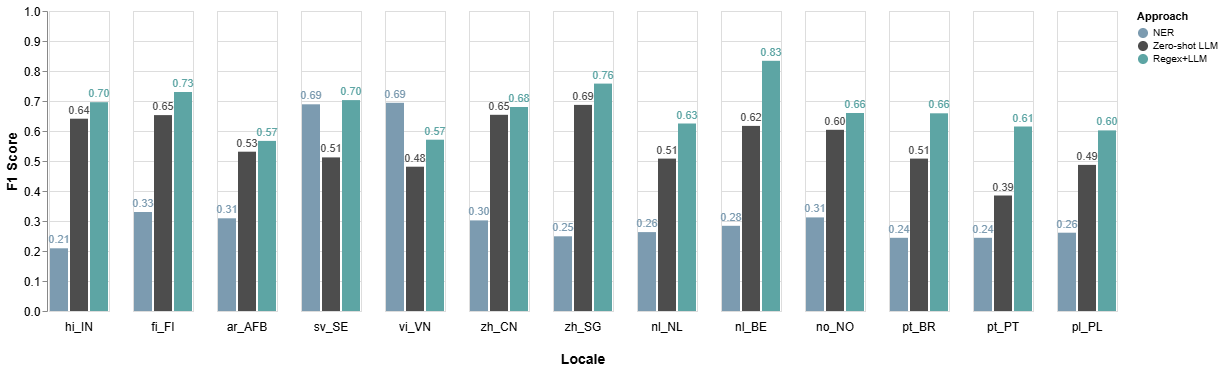}
    \caption{F1 Scores by Approach and Locale}
    \label{fig:f1-approach}
    \vspace{-7mm}
\end{figure*}


The evaluation dataset was created by experts, who authored text samples and injected synthetic PII across six domains (Finance, Travel, Healthcare, IT, CPG, Media). This allows for a control over entity types and distributions. Text length varied from short (<21 words), to medium (21–240 words), large (240–1000 words), and extra-large (up to 4500 words), and were uniformly selected to ensure robustness. We use the \texttt{nervaluate} library with an \textbf{Exact Evaluation Method}, which requires a predicted entity's character span to match the gold span exactly to be counted as correct. Since our pipeline resolves label ambiguity at a later stage, we focused on evaluating span accuracy independently. This rigorous method emphasizes precise boundary detection, which is critical for PII redaction tasks. The primary evaluation metrics used across all comparisons were Accuracy, Precision, Recall, and F1 Score for label imbalance.

\textbf{Comparative Results and Analysis:} Figure~\ref{fig:f1-approach} presents comparative results achieved by each approach across all locales (full results in Table~\ref{tab:ner_results}). In low-resource settings such as Polish (\textsc{pl\_PL}), \textbf{RECAP} (F1=0.60) achieves a 130.77\% relative improvement over the NER baseline (F1=0.26) and a 22.45\% improvement over the zero-shot LLM (F1=0.49). The \textbf{RECAP} framework consistently outperforms both traditional NER models and zero-shot LLMs across most locales, with \textsc{zh\_SG} and \textsc{nl\_BE} gaining F1 scores as high as 0.76 and 0.83, respectively. 


In sensitive PII detection tasks, recall is particularly important, as false negatives - missed detections - can lead to serious privacy breaches. While false positives may cause over-redaction, false negatives risk direct data exposure. \textbf{RECAP} achieves a weighted recall of 0.605, compared to 0.362 for NER (+67.13\%) and 0.437 for zero-shot LLMs (+38.44\%), demonstrating significantly stronger detection.


It is important to note that this is not a one-to-one comparison across approaches. Our architecture is explicitly designed to detect a fixed schema of 300+ predefined PII types across 13 locales. While LLMs were applied to detect this full set, their inherent variability often resulted in inconsistent formatting and hallucinated entities outside the defined set. In contrast, NER baselines are restricted to a narrow set of generic entity types, and do not encompass the broader set of sensitive identifiers commonly required in PII detection tasks. As such, while we report unified F1 scores, these results reflect fundamentally different label scopes and should be interpreted accordingly.
\vspace{-4mm}
\section{Conclusion}
\vspace{-3mm}

We presented \textbf{RECAP}, a hybrid PII detection architecture that combines regex patterns with prompt-based LLMs. \textbf{RECAP} addresses structural variation and low-resource challenges by leveraging rule-based precision and LLM-based contextual reasoning. Its modular design enables per-entity and per-locale customization without retraining. Benchmarked against a fine-tuned NER system and a zero-shot LLM, \textbf{RECAP} outperforms both across 13 diverse locales, particularly on complex and region-specific entities.
\vspace{-4mm} 
\section{Limitations and Future Work}
\vspace{-3mm}

\textbf{Limitations:} (1) Use of different NER models per locale limits consistency in cross-locale baseline comparisons; (2) Reliance on a single LLM (GPT-4o); other models may offer complementary strengths; (3) Synthetic benchmark data may not capture the full complexity of real-user text; (4) Label set mismatch between broad PII types (300+) and narrow NER types (e.g., PER, ORG) complicates direct evaluation; (5) Domain coverage in our work (6 domains) may not represent all production text types (e.g., legal, social media).

\textbf{Future Work:} (1) Investigate RL for automatic prompt optimization per label/locale; (2) Design perturbation-based evaluation for robustness testing; (3) Explore LLMs for automated regex generation~\citep{ye2020sketch}; (4) Apply knowledge distillation for on-device inference; (5) Develop active learning with expert feedback to refine regex and models.

\clearpage
\bibliographystyle{unsrtnat}
\bibliography{Auto_PII_Bibliography}

\clearpage

\appendix
\section{Appendix}

\begin{figure*}[htbp]
\centering
\begin{tcolorbox}[title=NER Model Codes, 
    coltitle=white, 
    colback=white, 
    colframe=black, 
    colbacktitle=black, 
    fonttitle=\bfseries, 
    boxrule=0.5pt,
    width=\textwidth
]
\begin{adjustbox}{minipage=1.1\linewidth}
\ttfamily
\begin{tabbing}
\hspace{2em} \= \kill
NERd: Davlan/distilbert-base-multilingual-cased-ner-hrl \citep{distilbert_base_multilingual_cased_ner_hrl} \\
NERk: KB/bert-base-swedish-cased-ner \citep{bert_base_swedish_cased_ner} \\
NERn: NlpHUST/ner-vietnamese-electra-base \citep{ner_vietnamese_electra_base} \\
NERj: julian-schelb/roberta-ner-multilingual \citep{roberta_ner_multilingual} \\
NERb: Babelscape/wikineural-multilingual-ner \citep{wikineural_multilingual_ner}
\end{tabbing}
\end{adjustbox}
\end{tcolorbox}
\caption{Pretrained model codes used for multilingual NER.}
\label{fig:ner-model-codes}
\end{figure*}

\begin{table}[htbp]
\centering
\setlength{\tabcolsep}{5pt}  
\renewcommand{\arraystretch}{1.05}  
\begin{tabular}{lc}
\toprule
\textbf{Locale} & \textbf{Samples} \\
\midrule
sv\_SE (Swedish -- Sweden) & 150 \\
vi\_VN (Vietnamese -- Vietnam) & 150+ \\
zh\_CN (Chinese -- China) & 105 \\
zh\_SG (Chinese -- Singapore) & 45 \\
pt\_BR (Portuguese -- Brazil) & 110+ \\
pt\_PT (Portuguese -- Portugal) & 45 \\
pl\_PL (Polish -- Poland) & 150+ \\
hi\_IN (Hindi -- India) & 150+ \\
fi\_FI (Finnish -- Finland) & 150 \\
ar\_AE (Arabic -- UAE) & 150 \\
nl\_NL (Dutch -- Netherlands) & 105 \\
nl\_BE (Dutch -- Belgium) & 45 \\
no\_NO (Norwegian -- Norway) & 150 \\
\bottomrule
\end{tabular}
\vspace{2mm}
\caption{Benchmark Sample Count by Locale}
\label{tab:locale_samples}
\end{table}


\begin{table*}[htbp]
\centering
\begin{tabular}{lcccccc}
\hline
\textbf{Locale} & \textbf{Phase I} & \textbf{Phase II} & \textbf{Phase III} & \textbf{$\Delta$ (1$\to$2)} & \textbf{$\Delta$ (2$\to$3)} & \textbf{$\Delta$ (1$\to$3)} \\
\hline
sv\_SE & 0.396 & 0.614 & 0.703 & 55.05\% & 14.50\% & 77.53\% \\
vi\_VN & 0.468 & 0.539 & 0.571 & 15.17\% & 5.94\% & 22.01\% \\
zh\_CN & 0.594 & 0.632 & 0.680 & 6.40\% & 7.60\% & 14.48\% \\
zh\_SG & 0.590 & 0.742 & 0.758 & 25.76\% & 2.16\% & 28.48\% \\
nl\_NL & 0.582 & 0.597 & 0.625 & 2.58\% & 4.69\% & 7.39\% \\
nl\_BE & 0.836 & 0.785 & 0.834 & -6.10\% & 6.24\% & -0.24\% \\
no\_NO & 0.583 & 0.664 & 0.660 & 13.89\% & -0.60\% & 13.21\% \\
hi\_IN & 0.486 & 0.503 & 0.696 & 3.50\% & 38.37\% & 43.21\% \\
fi\_FI & 0.573 & 0.592 & 0.730 & 3.32\% & 23.31\% & 27.40\% \\
ar\_AFB & 0.463 & 0.479 & 0.567 & 3.46\% & 18.37\% & 22.46\% \\
pt\_BR & 0.446 & 0.547 & 0.659 & 22.65\% & 20.48\% & 47.76\% \\
pt\_PT & 0.511 & 0.593 & 0.615 & 16.05\% & 3.71\% & 20.35\% \\
pl\_PL & 0.428 & 0.583 & 0.602 & 36.22\% & 3.26\% & 40.65\% \\
\hline
\end{tabular}
\caption{Locale Performance Across Three Phases by F1 Score}
\label{tab:locale_iterations}
\end{table*}

\begin{table*}[htbp]
\centering
\begin{tabular}{llcccccccc}
\toprule
\textbf{Locale} & \textbf{Approach} & \textbf{Accuracy} & \textbf{Precision} & \textbf{Recall} & \textbf{F1 Score} & \textbf{TP} & \textbf{FP} & \textbf{TN} & \textbf{FN} \\
\midrule
hi\_IN & NERd & 0.328 & 0.194 & 0.227 & 0.209 & 59 & 245 & 159 & 201 \\
hi\_IN & Zero-shot LLM & 0.472 & 0.801 & 0.534 & 0.641 & 310 & 77 & 0 & 22 \\
hi\_IN & RECAP & 0.534 & 0.781 & 0.628 & \textbf{0.696} & 364 & 102 & 0 & 216 \\
\midrule
fi\_FI & NERd & 0.347 & 0.431 & 0.268 & 0.330 & 166 & 219 & 192 & 454 \\
fi\_FI & Zero-shot LLM & 0.485 & 0.830 & 0.538 & 0.653 & 534 & 109 & 0 & 458 \\
fi\_FI & RECAP & 0.574 & 0.744 & 0.716 & \textbf{0.730} & 710 & 244 & 0 & 282 \\
\midrule
ar\_AE & NERd & 0.420 & 0.474 & 0.229 & 0.309 & 267 & 296 & 622 & 897 \\
ar\_AE & Zero-shot LLM & 0.362 & 0.667 & 0.442 & 0.531 & 886 & 443 & 0 & 1120 \\
ar\_AE & RECAP & 0.396 & 0.610 & 0.537 & \textbf{0.567} & 1078 & 736 & 0 & 928 \\
\midrule
sv\_SE & NERk & 0.705 & 0.599 & 0.810 & 0.689 & 205 & 137 & 237 & 48 \\
sv\_SE & Zero-shot LLM & 0.344 & 0.720 & 0.397 & 0.512 & 641 & 249 & 0 & 974 \\
sv\_SE & RECAP & 0.542 & 0.714 & 0.692 & \textbf{0.703} & 1118 & 447 & 0 & 497 \\
\midrule
vi\_VN & NERn & 0.737 & 0.602 & 0.818 & 0.694 & 198 & 131 & 292 & 44 \\
vi\_VN & Zero-shot LLM & 0.317 & 0.751 & 0.354 & 0.481 & 796 & 264 & 0 & 1454 \\
vi\_VN & RECAP & 0.400 & 0.772 & 0.453 & 0.571 & 1020 & 302 & 0 & 1230 \\
\midrule
zh\_CN & NERj & 0.385 & 0.446 & 0.228 & 0.302 & 120 & 149 & 228 & 406 \\
zh\_CN & Zero-shot LLM & 0.486 & 0.811 & 0.549 & 0.654 & 625 & 146 & 0 & 514 \\
zh\_CN & RECAP & 0.515 & 0.652 & 0.709 & \textbf{0.680} & 808 & 431 & 0 & 331 \\
\midrule
zh\_SG & NERj & 0.523 & 0.449 & 0.172 & 0.249 & 31 & 38 & 174 & 149 \\
zh\_SG & Zero-shot LLM & 0.523 & 0.799 & 0.603 & 0.687 & 279 & 70 & 0 & 184 \\
zh\_SG & RECAP & 0.610 & 0.753 & 0.762 & \textbf{0.758} & 353 & 116 & 0 & 110 \\
\midrule
nl\_NL & NERj & 0.280 & 0.238 & 0.294 & 0.263 & 160 & 511 & 188 & 385 \\
nl\_NL & Zero-shot LLM & 0.341 & 0.878 & 0.358 & 0.508 & 473 & 66 & 0 & 849 \\
nl\_NL & RECAP & 0.456 & 0.790 & 0.517 & \textbf{0.625} & 684 & 182 & 3 & 638 \\
\midrule
nl\_BE & NERj & 0.364 & 0.243 & 0.342 & 0.284 & 50 & 156 & 94 & 96 \\
nl\_BE & Zero-shot LLM & 0.446 & 0.840 & 0.487 & 0.617 & 194 & 37 & 0 & 204 \\
nl\_BE & RECAP & 0.715 & 0.887 & 0.786 & \textbf{0.834} & 313 & 40 & 0 & 85 \\
\midrule
no\_NO & NERj & 0.436 & 0.282 & 0.348 & 0.312 & 162 & 412 & 390 & 303 \\
no\_NO & Zero-shot LLM & 0.433 & 0.873 & 0.462 & 0.604 & 517 & 75 & 0 & 604 \\
no\_NO & RECAP & 0.493 & 0.785 & 0.570 & \textbf{0.660} & 638 & 175 & 0 & 481 \\
\midrule
pt\_BR & NERj & 0.309 & 0.293 & 0.210 & 0.244 & 136 & 328 & 240 & 513 \\
pt\_BR & Zero-shot LLM & 0.341 & 0.886 & 0.356 & 0.508 & 186 & 24 & 0 & 336 \\
pt\_BR & RECAP & 0.492 & 0.792 & 0.560 & \textbf{0.659} & 290 & 76 & 0 & 224 \\
\midrule
pt\_PT & NERj & 0.309 & 0.293 & 0.210 & 0.244 & 136 & 328 & 240 & 513 \\
pt\_PT & Zero-shot LLM & 0.239 & 0.878 & 0.247 & 0.385 & 158 & 22 & 0 & 482 \\
pt\_PT & RECAP & 0.444 & 0.689 & 0.555 & \textbf{0.615} & 354 & 160 & 0 & 284 \\
\midrule
pl\_PL & NERb & 0.276 & 0.394 & 0.195 & 0.261 & 133 & 208 & 154 & 550 \\
pl\_PL & Zero-shot LLM & 0.322 & 0.744 & 0.362 & 0.487 & 128 & 44 & 0 & 226 \\
pl\_PL & RECAP & 0.425 & 0.614 & 0.579 & \textbf{0.602} & 398 & 244 & 0 & 242 \\
\midrule
All locales & NER* & 0.428 & 0.395 & 0.362 & 0.360 & - & - & - & - \\
All locales & Zero-shot LLM* & 0.391 & 0.795 & 0.437 & 0.558 & - & - & - & - \\
All locales & RECAP* & 0.492 & 0.729 & 0.605 & \textbf{0.657} & - & - & - & - \\
\bottomrule
\end{tabular}
\caption{PII Detection Performance Results by Locale and Approach}
\label{tab:ner_results}
\textbf{Note:} The last three rows (marked by *) represent weighted averages across all locales.
\end{table*}


\begin{table*}[htbp]
\centering
\begin{tabular}{llcccccccc}
\toprule
\textbf{Locale} & \textbf{Phase} & \textbf{Accuracy} & \textbf{Precision} & \textbf{Recall} & \textbf{F1 Score} & \textbf{TP} & \textbf{FP} & \textbf{TN} & \textbf{FN} \\
\midrule
sv\_SE & I & 0.247 & 0.438 & 0.362 & 0.396 & 584 & 749 & 0 & 1031 \\
sv\_SE & II & 0.443 & 0.582 & 0.650 & 0.614 & 1049 & 753 & 0 & 566 \\
sv\_SE & III & 0.542 & 0.714 & 0.692 & 0.703 & 1118 & 447 & 0 & 497 \\
\midrule
vi\_VN & I & 0.326 & 0.551 & 0.407 & 0.468 & 915 & 747 & 91 & 1335 \\
vi\_VN & II & 0.369 & 0.630 & 0.471 & 0.539 & 1059 & 622 & 0 & 1191 \\
vi\_VN & III & 0.400 & 0.772 & 0.453 & 0.571 & 1020 & 302 & 0 & 1230 \\
\midrule
zh\_CN & I & 0.481 & 0.545 & 0.654 & 0.594 & 745 & 623 & 199 & 394 \\
zh\_CN & II & 0.462 & 0.582 & 0.692 & 0.632 & 788 & 566 & 0 & 351 \\
zh\_CN & III & 0.515 & 0.652 & 0.709 & 0.680 & 808 & 431 & 0 & 331 \\
\midrule
zh\_SG & I & 0.487 & 0.573 & 0.607 & 0.590 & 281 & 209 & 90 & 182 \\
zh\_SG & II & 0.590 & 0.723 & 0.762 & 0.742 & 353 & 135 & 0 & 110 \\
zh\_SG & III & 0.610 & 0.753 & 0.762 & 0.758 & 353 & 116 & 0 & 110 \\
\midrule
nl\_NL & I & 0.434 & 0.652 & 0.526 & 0.582 & 695 & 371 & 69 & 627 \\
nl\_NL & II & 0.448 & 0.665 & 0.542 & 0.597 & 716 & 361 & 69 & 606 \\
nl\_NL & III & 0.456 & 0.790 & 0.517 & 0.625 & 684 & 182 & 3 & 638 \\
\midrule
nl\_BE & I & 0.718 & 0.909 & 0.774 & 0.836 & 308 & 31 & 0 & 90 \\
nl\_BE & II & 0.645 & 0.871 & 0.714 & 0.785 & 284 & 42 & 0 & 114 \\
nl\_BE & III & 0.715 & 0.887 & 0.786 & 0.834 & 313 & 40 & 0 & 85 \\
\midrule
no\_NO & I & 0.442 & 0.677 & 0.512 & 0.583 & 573 & 273 & 75 & 546 \\
no\_NO & II & 0.497 & 0.780 & 0.578 & 0.664 & 647 & 182 & 0 & 472 \\
no\_NO & III & 0.493 & 0.785 & 0.570 & 0.660 & 638 & 175 & 0 & 481 \\
\midrule
hi\_IN & I & 0.347 & 0.406 & 0.607 & 0.486 & 352 & 516 & 44 & 228 \\
hi\_IN & II & 0.336 & 0.426 & 0.614 & 0.503 & 356 & 479 & 0 & 224 \\
hi\_IN & III & 0.534 & 0.781 & 0.628 & 0.696 & 364 & 102 & 0 & 216 \\
\midrule
fi\_FI & I & 0.416 & 0.493 & 0.685 & 0.573 & 680 & 700 & 42 & 312 \\
fi\_FI & II & 0.421 & 0.515 & 0.698 & 0.592 & 692 & 652 & 0 & 300 \\
fi\_FI & III & 0.574 & 0.744 & 0.716 & 0.730 & 710 & 244 & 0 & 282 \\
\midrule
ar\_AE & I & 0.316 & 0.421 & 0.514 & 0.463 & 1032 & 1421 & 73 & 974 \\
ar\_AE & II & 0.315 & 0.442 & 0.522 & 0.479 & 1048 & 1323 & 0 & 958 \\
ar\_AE & III & 0.396 & 0.610 & 0.537 & 0.567 & 1078 & 736 & 0 & 928 \\
\midrule
pt\_BR & I & 0.320 & 0.453 & 0.439 & 0.446 & 214 & 258 & 36 & 274 \\
pt\_BR & II & 0.377 & 0.519 & 0.579 & 0.547 & 294 & 214 & 0 & 214 \\
pt\_BR & III & 0.492 & 0.792 & 0.560 & 0.659 & 290 & 76 & 0 & 224 \\
\midrule
pt\_PT & I & 0.366 & 0.536 & 0.489 & 0.511 & 312 & 270 & 32 & 326 \\
pt\_PT & II & 0.421 & 0.620 & 0.568 & 0.593 & 352 & 216 & 0 & 268 \\
pt\_PT & III & 0.444 & 0.689 & 0.555 & 0.615 & 354 & 160 & 0 & 284 \\
\midrule
pl\_PL & I & 0.297 & 0.416 & 0.440 & 0.428 & 308 & 432 & 40 & 394 \\
pl\_PL & II & 0.412 & 0.567 & 0.601 & 0.583 & 364 & 278 & 0 & 242 \\
pl\_PL & III & 0.425 & 0.614 & 0.579 & 0.602 & 398 & 244 & 0 & 242 \\
\midrule
All locales & I* & 0.372 & 0.516 & 0.522 & 0.511 & - & - & - & - \\
All locales & II* & 0.419 & 0.584 & 0.601 & 0.585 & - & - & - & - \\
All locales & III* & 0.492 & 0.729 & 0.605 & 0.657 & - & - & - & - \\
\bottomrule
\end{tabular}
\caption{PII Detection Performance Results by Locale and Phase}
\label{tab:ner_iteration_results}
\textbf{Note:} The last three rows (marked by *) represent weighted averages across all locales.
\end{table*}



\clearpage
\begin{lstlisting}[style=prompt,
  caption={High-level overview of Name-extraction prompt used in experiments},
  label={lst:name-prompt}]
{ 
System Prompt: You are a multilingual PII detection system. Your task is to detect and extract personal names from the input text based on the specified {locale}. Follow these rules: - Always consider locale-specific naming conventions. - Return names exactly as they appear in the text (including diacritics, prefixes, and original scripts). - If no names are found, return an empty list []. Locale-specific examples: - zh_CN / zh_SG: Extract Chinese names in original characters. - vi_VN: Vietnamese names follow the order [Family Name] [Middle Name] [Given Name]. - nl_NL / nl_BE: Dutch/Flemish names may include prefixes (e.g., "van", "de"). User Prompt: LOCALE: {locale} TEXT: "{text}" 
} 
\end{lstlisting}

\end{document}